\documentclass[lettersize,journal]{IEEEtran}
\usepackage{amsmath,amsfonts}
\usepackage{algorithmic}
\usepackage{algorithm}
\usepackage{array}
\usepackage[caption=false,font=normalsize,labelfont=sf,textfont=sf]{subfig}
\usepackage{textcomp}
\usepackage{stfloats}
\usepackage{url}
\usepackage{verbatim}
\usepackage{graphicx}
\usepackage{multirow}
\usepackage{amssymb}
\usepackage{xcolor}
\usepackage[capitalize]{cleveref}
\crefname{section}{Sec.}{Secs.}
\Crefname{section}{Section}{Sections}
\Crefname{table}{Table}{Tables}
\crefname{table}{Tab.}{Tabs.}
\crefname{equation}{Eq.}{Eqs.}

\usepackage{cite}
\begin{document}

\title{Retrieval-augmented Few-shot Medical Image Segmentation with Foundation Models}

\author{Lin Zhao, Xiao Chen, Eric Z. Chen, Yikang Liu, Terrence Chen, Shanhui Sun
\thanks{L. Zhao, Y. Liu, X. Chen, E. Chen,  T. Chen and S. Sun are with United Imaging Intelligence, 65 Blue Sky Drive, Burlington, MA 01803, USA. (e-mail: shanhui.sun@uii-ai.com)}}


\maketitle

\begin{abstract}
Medical image segmentation is crucial for clinical decision-making, but the scarcity of annotated data presents significant challenges. Few-shot segmentation (FSS) methods show promise but often require training on the target domain and struggle to generalize across different modalities. Similarly, adapting foundation models like the Segment Anything Model (SAM) for medical imaging has limitations, including the need for finetuning and domain-specific adaptation. To address these issues, we propose a novel method that adapts DINOv2 and Segment Anything Model 2 (SAM 2) for retrieval-augmented few-shot medical image segmentation. Our approach uses DINOv2's feature as query to retrieve similar samples from limited annotated data, which are then encoded as memories and stored in memory bank. With the memory attention mechanism of SAM 2, the model leverages these memories as conditions to generate accurate segmentation of the target image. We evaluated our framework on three medical image segmentation tasks, demonstrating superior performance and generalizability across various modalities without the need for any retraining or finetuning. Overall, this method offers a practical and effective solution for few-shot medical image segmentation and holds significant potential as a valuable annotation tool in clinical applications.

\end{abstract}


\section{Introduction}

\IEEEPARstart{M}{edical} image segmentation plays a critical role in diagnosis, treatment planning,enabling precise analysis for various clinical applications.~\cite{rogowska2000overview,siddique2021u,azad2024medical}. Accurate segmentation enables clinicians to precisely delineate anatomical structures and abnormalities within images, which is essential for making informed clinical decisions. However, unlike natural image segmentation, medical image segmentation faces unique challenges due to the limited availability of data and annotations~\cite{chi2020deep,cui2020unified,peng2021medical}. Obtaining precise annotations requires the expertise of trained medical professionals and is time-consuming and labor-intensive, making it difficult to gather enough labeled samples for supervised training. In practice, while acquiring a large number of annotations is often impractical, obtaining a smaller set, such as a few dozen, is more feasible. Maximizing the utility of these limited samples to create reliable segmentation masks without compromising the accuracy is a pressing challenge in the field.


\begin{figure}[t]
\begin{center}
\includegraphics[width=1\linewidth]{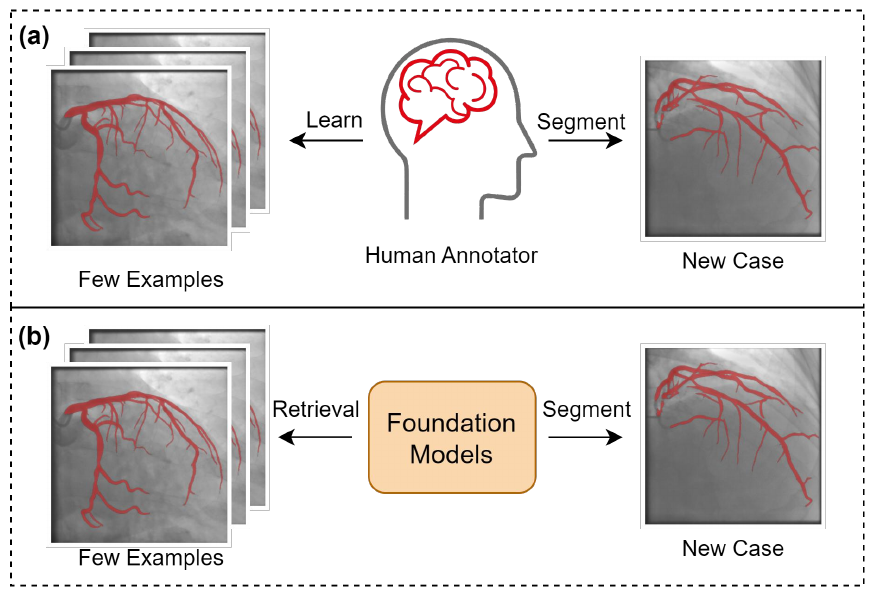}
\end{center}
\caption{(a) Illustration of how human annotators learn to segment medical images: By studying a few annotated examples, the human annotator can effectively apply the learned knowledge to segment a new, unseen case.(b) Representation of our proposed model's process: Retrieving the contextual and anatomical information from similar annotated example to guide the foundation models to perform the segmentation for new case without any retraining or finetuning.}
\label{fig:fig1}
\end{figure}

Few-shot segmentation (FSS) offers promising potential to address the aforementioned challenges in the medical imaging domain, particularly the scarcity of labeled data~\cite{wang2019panet}. Various approaches have been developed to enable models to learn from a limited number of annotated samples, such as ALPNet~\cite{ouyang2020self}, ADNet~\cite{hansen2022anomaly},Q-Net~\cite{shen2023q}, and CAT-Net~\cite{lin2023few}. Despite their effectiveness and success, these methods have certain limitations. They often require self-supervised pretraining on the target domain~\cite{ouyang2020self,hansen2022anomaly}, or training with known classes before being applied to new classes~\cite{lin2023few}, limiting their ability to generalize across different image modalities. Additionally, most of these methods employ a 1-way 1-shot setting. While this setting can be valuable in research, acquiring a small number of annotated samples is feasible and aligns better with clinical practices. Furthermore, the compromised performance of these methods may not match that of supervised methods trained on large datasets, limiting their ability to reduce the annotation burden. These limitations restrict the practical application and effectiveness of these models in diverse clinical scenarios.

Foundation models, pretrained on large-scale datasets across various domains, have demonstrated strong few-shot and zero-shot capabilities. In the realm of image segmentation, the Segment-Anything Model (SAM) has shown remarkable zero-shot generalization when driven by prompts, suggesting its potential for FSS in medical images. However, SAM was pretrained on natural images and underperforms in specialized areas like medical imaging, leading to various adaptations. These adaptations include constructing large-scale medical image datasets for fine-tuning SAM~\cite{ma2024segment}, using adapters to tailor SAM for domain-specific needs~\cite{wu2023medical}, and extending SAM from 2D to 3D medical image segmentation~\cite{lei2023medlsam,bui2023sam3d}. While these methods have improved performance compared to fully supervised models like U-Net~\cite{ronneberger2015u,isensee2021nnu}, they still require fine-tuning and retraining on specific domains or image modalities. Given a small amount of annotated data, fine-tuning and retraining may be ineffective. Intuitively, we could leverage these small annotated samples as prompts to guide the model in segmentation tasks, similar to how prompts guide large language models (LLMs). However, current methods still rely on prompts like points, boxes, or masks, which do not fully exploit the potential of the available annotated data.

Recently, the Segment Anything Model 2 (SAM 2) introduced enhanced capabilities for image segmentation and extended promptable segmentation to video applications~\cite{ravi2024sam}. A key modification in SAM 2 for video segmentation is the memory-driven approach which leverages temporal information by encoding segmentations from previous video frames as memories and condition on those memories to produce the segmentation of subsequent frames. Such ability naturally adapts to few-shot medical image segmentation: instead of integrating information from previous frames, we can retrieval the contextual and anatomical information from similar cases to guide the segmentation process, as the Retrieval-Augmented Generation (RAG) approach used in LLMs~\cite{lewis2020retrieval,edge2024local}. This mirrors how humans learn to perform segmentation tasks: even without prior knowledge, they can observe and study a few examples, quickly understand the process, and then apply it effectively (\Cref{fig:fig1}).

Building on this intuition, we propose a novel framework that adapts DINOv2~\cite{oquab2023dinov2} and SAM 2~\cite{ravi2024sam} for few-shot medical image segmentation. The framework introduces a retrieval module that uses DINOv2 to generate embeddings and retrieve similar samples from limited annotated data. The segmentation network, adapted from SAM 2, encodes these retrieved images and masks as memories stored in a memory bank. The memory attention mechanism then integrates relevant features and anatomical structures from these stored memories, enhancing the input image features for the decoder which generates the final segmentation mask with greater precision and accuracy. Notably, our framework directly utilizes the structure and pre-trained weights without any retraining or finetuning in the target domain. When applied to a new domain, only the samples in the retrieval database need to be updated. We evaluate the proposed framework on three medical image segmentation datasets, including public ACDC dataset~\cite{bernard2018deep}, and in-house CMR T1-Map, and Fluoroscopy Image datasets. Building on these foundation models, our method demonstrates superior performance and generalizes well across different modalities and contrasts. Compared with other few-shot medical image segmentation methods, our method achieve state-of-the-art performance without any retraining or fine-tuning in ABD-MRI dataset~\cite{kavur2021chaos}. Our contributions in this work are summarized as follows:

\begin{itemize}
    \item We introduce a novel framework that integrates retrieval augmentation and memory-driven approach, leveraging foundation models like DINOv2 and SAM 2 for accurate few-shot medical image segmentation with limited annotated data.

    \item We present an effective method to adapt retrieval-augmented techniques for querying relevant annotated samples in medical image segmentation. To the best of our knowledge, this is the first work to apply retrieval-augmented techniques to image segmentation.
    
    \item Our framework demonstrates exceptional adaptability across various imaging modalities and contrasts without requiring retraining, highlighting its potential as a valuable annotation tool in diverse clinical scenarios.

\end{itemize}

\section{Related Works}

\subsection{Foundation Models}
Large-scale language, vision, and multimodal foundation models such as GPTs~\cite{radford2018improving, achiam2023gpt, dubey2024llama}, DINO (self-Distillation with No labels)~\cite{caron2021emerging, oquab2023dinov2}, Contrastive Language-Image Pretraining (CLIP)~\cite{radford2021learning}, and Large Language and Vision Assistant (LLaVA)~\cite{liu2024visual} have dramatically transformed the artificial intelligence landscape. These models significantly enhance zero-shot and few-shot capabilities across natural language processing, image recognition, and multimodal applications. In image domain, DINOv2 has demonstrated exceptional proficiency in learning high-level semantic information, facilitating a variety of downstream tasks such as image classification, semantic segmentation, image retrieval, video tracking~\cite{tumanyan2024dino}, and feature matching~\cite{jiang2024omniglue}. Another notable development is the Segment Anything Model (SAM)~\cite{kirillov2023segment}, a foundation model pretrained on a large-scale segmentation dataset, has shown outstanding zero-shot generalization capabilities in promptable segmentation. However, SAM underperformed in some specific domains such as medical imaging~\cite{mazurowski2023segment} so various attempts have been made in fintuning and adapting it~\cite{wu2023medical,ma2024segment}. Recently, Segment Anything Model 2 (SAM 2) has brought enhanced capabilities for segmenting images and extending promptable segmentation to video applications~\cite{ravi2024sam}. In this paper, we leverage the semantic learning of DINOv2 for retrieval augmentation and perform the few-shot segmentation by adapting the video segmentation capabilities of SAM 2. This integration allows us to retrieval similar images with masks as prompts to guide segmentation for medical images without further training/finetuning as in previous works~\cite{wu2023medical,ma2024segment}.

\subsection{Retrieval-augmented Techniques}
Retrieval-Augmented Generation (RAG) in natural language processing combines the strengths of retrieval-based systems and generative models~\cite{lewis2020retrieval,edge2024local}. RAG begins by retrieving relevant information from an external knowledge source based on the input query, providing contextual information that enhances the accuracy, informativeness, and contextual grounding of responses generated by large language models (LLMs). In the image domain, retrieval-augmented techniques have also been explored in various areas. For example, the Retrieval-Augmented Diffusion Model (RDM) was proposed for image synthesis, where the generative model is conditioned on informative samples retrieved from a database~\cite{blattmann2022retrieval}. Similarly, Retrieval-Augmented Classification (RAC) integrates an explicit retrieval module into standard image classification pipelines to improve long-tail visual recognition~\cite{long2022retrieval}. Additionally, REtrieval-Augmented CusTomization (REACT) leverages relevant image-text pairs from a vast web-scale database to customize visual models for specific target domains, achieving significant performance across tasks while minimizing the need for extensive retraining~\cite{liu2023learning}. In this work, we extend the retrieval strategy to the image segmentation task, designing a retrieval module that queries an external database for contextual and anatomical information, which is then used to guide the SAM 2 model in few-shot medical image segmentation without requiring any retraining.

\subsection{Few-shot Medical Image Segmentation}
Few-shot segmentation (FSS) has been widely explored in the medical imaging field, where data and annotations are usually scarce. Typically, FSS frameworks follow the Prototype Alignment Network (PANet) approach, which learns class-specific prototype representations from a few support images and then performs segmentation on the query images by matching each pixel to the learned prototypes~\cite{wang2019panet}. For the FSS in medical imaging domain, ALPNet enhances PANet by introducing superpixel-based self-supervised learning, eliminating the need for manual annotations, and by incorporating an adaptive local prototype pooling module to address the foreground-background imbalance in medical image segmentation~\cite{ouyang2020self}. The Anomaly Detection-inspired Network (ADNet) takes a different approach by relying on a single foreground prototype to compute anomaly scores for all query pixels, with segmentation achieved by thresholding these scores using a learned threshold~\cite{hansen2022anomaly}. Q-Net improves the ADNet by adding a query-informed threshold adaptation module and a query-informed prototype refinement module~\cite{shen2023q}. 
CAT-Net, which leverages a cross-masked attention transformer, enhances the FSS by mining the correlations between the support and query images and restricts the focus to relevant foreground information~\cite{lin2023few}. Despite the success of these FSS methods, they still require self-supervised pretraining or training on the base classes before applying to new classes in the same domain. This limitation restricts their practicality in real-world applications. Our approach leverage the foundation models to perform the FSS without any training in new domain, enabling more flexible and efficient segmentation while achieving state-of-the-art performance.
\section{Methods}

\begin{figure*}[t]
\begin{center}
\includegraphics[width=1\linewidth]{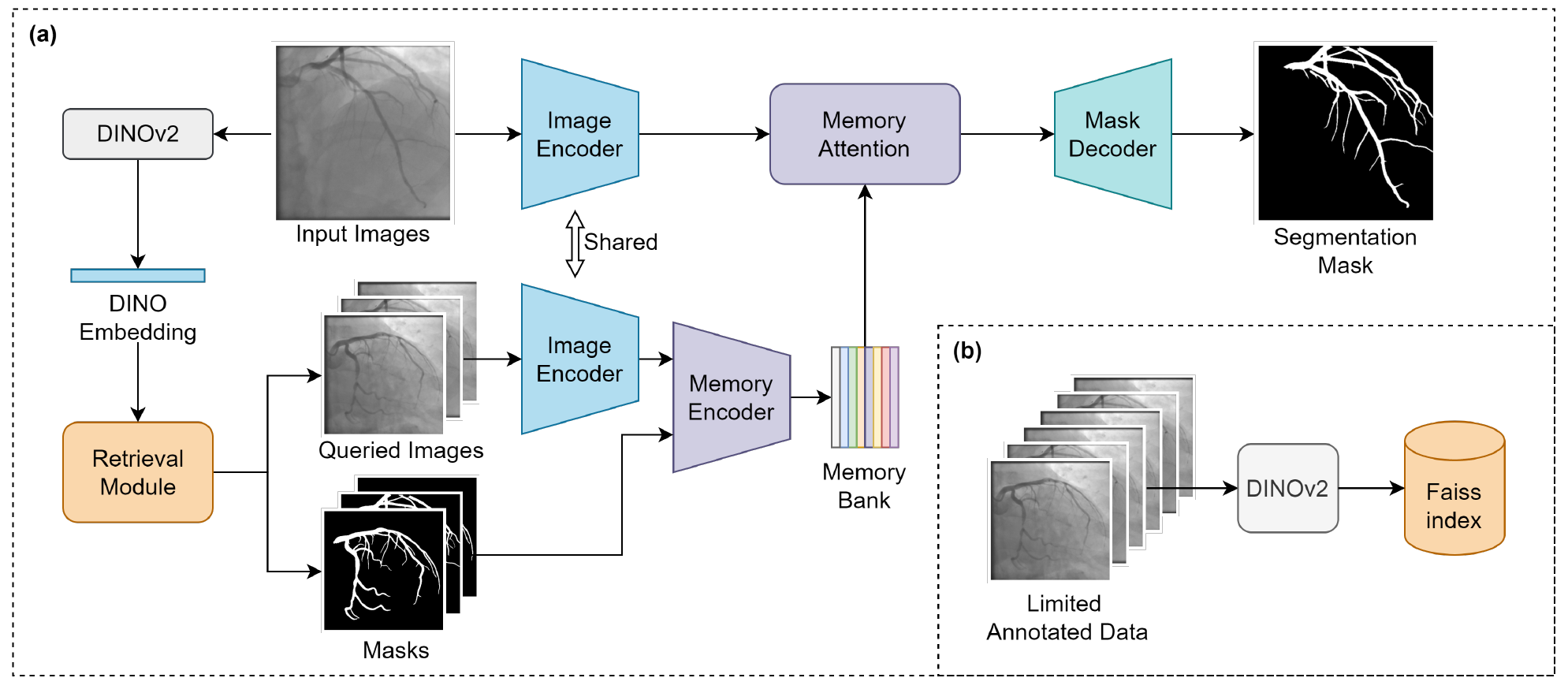}
\end{center}
\caption{Overview of Retrieval-Augmented Few-Shot Medical Image Segmentation Workflow: (a) The main segmentation pipeline starts with input images processed through DINOv2 for dino embedding, followed by querying similar images and corresponding masks which are encoded and stored in a memory bank. The memory attention mechanism integrates the information from memory bank to assist the mask decoder in generating the final segmentation mask. (b) The process of indexing limited annotated data using DINOv2 and Faiss, enabling efficient retrieval of relevant images to enhance segmentation accuracy.}
\label{fig:fig2}
\end{figure*}

\subsection{Preliminaries of SAM 2 Model}

SAM 2 (Segment Anything Model 2) is designed for both image and video segmentation, featuring several key components. It utilizes a hierarchical image encoder, pre-trained with MAE and Hiera, which captures multiscale image features for robust image representation. The memory attention mechanism in SAM 2 dynamically integrates information from past frames, leveraging self-attention and cross-attention within transformer blocks to condition current frame features on past data. This mechanism works in tandem with a memory bank that stores encoded memories of previous frames, allowing SAM 2 to recall and utilize temporal context effectively. Additionally, SAM 2's mask decoder and prompt encoder enhance segmentation precision by incorporating high-resolution details and handling ambiguous prompts through predictive modeling.

The memory-driven approach of SAM 2 for video segmentation is both inspiring and intuitive for our medical image segmentation task. Its ability to recall and integrate relevant past information naturally adapts to few-shot segmentation scenarios. Instead of integrating information from previous frames, we can retrieve the contextual and anatomical information from similar cases to guide the segmentation process. This mirrors how humans learn to perform segmentation tasks: even without prior knowledge, by observing and understanding a few examples, they can quickly grasp the task and execute it accurately. Our framework builds on these intuitions; in \Cref{sec:retrieval}, we detail the retrieval module that retrieval the similar cases for supporting the segmenation process, and in \Cref{sec:fss}, we introduce our few-shot segmentation network which is adapted from the SAM 2 model.

\subsection{Retrieval Module}
\label{sec:retrieval}
The Retrieval Module is a critical component of our segmentation framework, designed to augment segmentation by efficiently retrieving relevant annotated examples from a database built from limited samples. The module begins by utilizing DINOv2~\cite{oquab2023dinov2} to extract high-dimensional embeddings from a set of annotated images. These embeddings capture the semantic features of the images and are then indexed using FAISS (Facebook AI Similarity Search)~\cite{douze2024faiss} (\Cref{fig:fig2}(b)) which is optimized for efficient similarity search of dense vectors.

Once the FAISS index of annotated images is constructed, the Retrieval Module leverages DINOv2-generated embeddings of the input image to query the index and identify the most similar images. These queried images, along with their associated segmentation masks, are then passed to the next stage of the pipeline, where they are encoded and stored in a memory bank. This process ensures that the model can access relevant contextual and anatomical information during segmentation, guiding the model to make more accurate and informed segmentation decisions.

\subsection{Few-shot Segmentation Framework}
\label{sec:fss}
Our few-shot segmentation framework, as shown in \Cref{fig:fig2} (a), is designed to perform medical image segmentation with limited annotated data by leveraging a retrieval-augmented system. This framework is adapted from the Segment Anything Model 2 (SAM 2)~\cite{ravi2024sam} and operates without requiring additional training, making it highly adaptable and efficient for various medical imaging tasks.

Given an input image, it is first processed through DINOv2 to generate embedding that capture its semantic information. This embedding is then fed into the Retrieval Module, as discussed in \Cref{sec:retrieval}, to retrieve the most similar images from a pre-built FAISS index. The retrieved images are passed through a pre-trained hierarchical image encoder to obtain the image embeddings, specifically a MAE-pretrained Hiera model which is designed to handle multiscale features for robust image representation. This hierarchical structure allows the framework to extract detailed and context-rich embeddings at multiple scales, which is crucial for the segmentation task. Once the image embeddings of the retrieved images are obtained, they are input into the memory encoder along with their associated segmentation masks. The memory encoder downsamples the segmentation masks and fuses them with the image embeddings obtained from the image encoder, generating compact and efficient memory representations. These encoded memories are then stored in a memory bank for the memory attention operation.

During the segmentation process, the input image is processed through the image encoder to obtain its embeddings. The model then utilizes the memory attention mechanism from SAM 2, which dynamically integrates information from the memory bank. This memory attention module consists of stacked transformer blocks that first apply self-attention to the current image features, followed by cross-attention to the stored memories. The enriched features generated by this process are then fed into the mask decoder. The mask decoder in SAM 2 is similar to the one in the original SAM, but with a key enhancement: skip connections from the hierarchical image encoder bypass the memory attention module. These skip connections incorporate high-resolution information directly into the mask decoding process, enhancing the accuracy and detail of the final segmentation output. This use of skip connections is particularly important in medical image segmentation, as it mirrors the architecture of U-Net, which is well-suited for capturing fine-grained details in medical images. More details refer to ~\cite{ravi2024sam}.

It is important to note that our framework directly adapts the image encoder, memory encoder, memory attention, and mask decoder from SAM 2 without requiring any additional retraining or pretraining. During segmentation, our framework operates without the need for external prompts for the mask decoder. Instead, the segmentation is guided by the rich contextual information stored in the memory bank, which includes encoded features from similar images. This memory-driven approach allows the framework to accurately segment target structures by leveraging past experiences, making it highly efficient and effective even with limited annotated data.

\section{Experiments}

\subsection{Datasets}
We employ 3 different medical image segmentation datasets to validate the proposed method.

\subsubsection{ACDC Dataset} Automated Cardiac Diagnosis Challenge (ACDC) dataset is a benchmark in cardiac MRI segmentation, containing images from 100 patients across five clinical conditions. It includes 2D short-axis cine-MRI sequences at end-diastolic (ED) and end-systolic (ES) phases, with manual segmentations of the left ventricle (LV), myocardium (Myo), and right ventricle(RV). To construct the image retrieval database, we randomly selected 3 subjects from the 100 patients and extracted a total of 50 axial slices from both the ED and ES phases. The same procedure was applied to create the test dataset by randomly selecting another 3 subjects. For comparison, the supervised training on the ACDC dataset uses the same test dataset but includes a training set with 1808 axial slices from 70 subjects.

\subsubsection{CMR T1-Map Dataset}
The CMR T1-Map dataset is dedicated to segmenting the myocardium (Myo) in T1 mapping CMR images, acquired under varying scanning protocols. To create a representative dataset, we randomly selected 3 subjects from a pool of 163 patients, extracting a total of 50 axial slices for retrieval. Similarly, an additional 3 subjects with 50 slices were selected for the test dataset. For comparison, the supervised training uses the same test dataset and a training dataset with an 402 axial slices from 25 subjects.

\subsubsection{Fluoroscopy Image Dataset}
The Fluoroscopy Image dataset is designed for the segmentation of the left coronary artery (LCA). It includes data from 9 subjects with annotations, with 5 subjects used to construct the retrieval database and 4 subjects designated for the test dataset. For each subject, we selected 10 consecutive frames from two different positional angles, resulting in a total of 20 frames per subject. This setup yields 100 frames for the retrieval database and 80 frames for the test dataset. For supervised methods, we sampled an additional 80 frames for each subject, resulting in a training set of 500 frames.

\subsection{Implementation Details}
The DINOv2 model we utilize for produce embeddings for retrieval is DINOv2 ViT-14-small with registers~\cite{darcet2023vitneedreg}, which yield a 384-dimensional vector for each image. For the SAM 2 model, we utilize the one build on Hiera-large~\cite{ryali2023hiera}. The retrieval module is implemented with Faiss library (\url{https://ai.meta.com/tools/faiss/}) with squared Euclidean (L2) distance~\cite{douze2024faiss}. In addition, we normalize the DINOv2 embeddings both when constructing the index and during retrieval.

\subsection{Segmentation Results on Three Datasets}
\begin{table*}[ht!]
\centering
\caption{Segmentation performance comparison using Dice similarity coefficient (DSC) on ACDC, CMR T1-MAP, and Fluoroscopy Image datasets. Abbreviations: RV (right ventricle), Myo (myocardium), LV (left ventricle).}
\begin{tabular}{lccccc}
\hline
\multirow{2}{*}{Methods} & \multicolumn{3}{c}{ACDC} &CMR T1-MAP & \multirow{2}{*}{Fluoroscopy Image} \\
\cline{2-4} \cline{5-5}
\textbf{} & RV & Myo & LV & Myo &  \\
\hline
SAM 2 (1 Pos Point) & 0.4146 & 0.4565 & 0.6612 & 0.4483 & 0.2788 \\
SAM 2 (2 Pos Points) & 0.4268 & 0.4604 & 0.6610 & 0.4436 & 0.2736 \\
SAM 2 (1 Pos\&Neg Points) & 0.4441 & 0.4794 & 0.6952 & 0.4215 & 0.2592 \\
SAM 2 (2 Pos\&Neg Points) & 0.4133 & 0.5004 & 0.6801 & 0.4357 & 0.3477 \\
Ours & \textbf{0.6729} & \textbf{0.7757} & \textbf{0.8472} & \textbf{0.8238} & \textbf{0.7029} \\
\hline
\end{tabular}
\label{tab:tab1}

\end{table*}
In this subsection, we present a comparative analysis of our model and the SAM 2 model using different point prompts. Table \Cref{tab:tab1} demonstrate the segmentation performance across three datasets: ACDC, CMR T1-MAP, and Fluoroscopy Image, evaluated using the Dice similarity coefficient (DSC). Our approach consistently outperforms the SAM 2 model across all datasets and point prompt configurations. For the ACDC dataset, our method achieves DSC scores of 0.6729, 0.7757, and 0.8472 for the right ventricle (RV), myocardium (Myo), and left ventricle (LV), respectively, demonstrating a remarkble improvement, particularly in the Myo and LV classes. In the CMR T1-MAP and Fluoroscopy Image datasets, our model also shows superior performance with DSC scores of 0.8238 and 0.8431, respectively. This consistent improvements in segmentation accuracy suggest the robustness and effectiveness of our method for few-shot medical image segmentation, especially in cases where the SAM 2 model with prompts falls short.

\begin{figure*}[t]
\begin{center}
\includegraphics[width=1\linewidth]{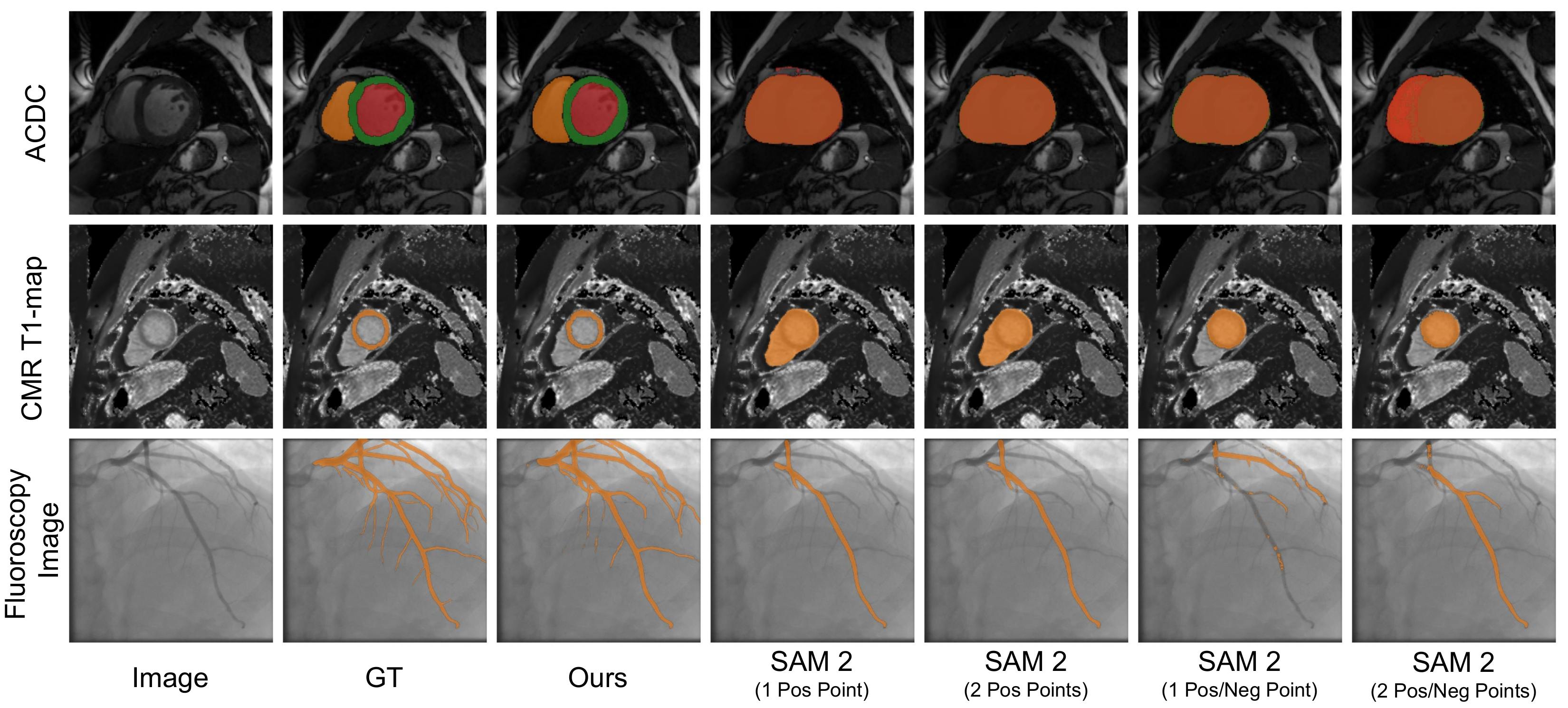}
\end{center}
\caption{Qualitative comparison of segmentation performance on ACDC, CMR T1-MAP, and Fluoroscopy Image datasets from a randomly selected sample for each dataset, respectively.}
\label{fig:fig3}
\end{figure*}

\Cref{fig:fig3} presents a qualitative comparison of segmentation performance across three datasets. In the ACDC dataset, our method closely matches the ground truth, accurately segmenting both the myocardium and ventricles. In contrast, the SAM 2 model with point prompts tends to over-segment the RV, Myo, and LV, particularly the ring-shaped myocardium. Similarly, in the CMR T1-MAP dataset, our method delivers more precise segmentation, while the SAM 2 variants often over-segment the myocardial region, incorrectly including parts of the LV and RV. For the Fluoroscopy Image dataset, our method achieves more refined and continuous vessel segmentation, whereas the SAM 2 model produces incomplete and less accurate results, missing several main and side branches. Overall, this figure clearly demonstrates the superiority of our method in achieving more accurate and reliable segmentations across diverse medical imaging modalities.

\subsection{Comparison with Supervised Methods}

We compare the segmentation performance of our method with fully-supervised methods such as U-Net~\cite{ronneberger2015u} and SwinUNETR~\cite{hatamizadeh2021swin} on three datasets, and present the results under both full and limited data settings in \Cref{tab:tab2}. When trained with full dataset, U-Net and SwinUNETR achieve high DSCs in ACDC and CMR T1-map datasets. However, in the limited sample setting with only 50 samples, the performance of these fully-supervised models drops, with U-Net and SwinUNETR showing markedly lower DSCs, especially for the right ventricle (RV) and myocardium (Myo) in ADCD dataset.

\begin{table*}[h!]
\centering
\caption{Comparison of segmentation performance on the ACDC, CMR T1-MAP, and Fluoroscopy Image datasets between fully-supervised methods and our approach.}
\begin{tabular}{llccccc}
\hline
\multirow{2}{*}{Settings} & \multirow{2}{*}{Methods} & \multicolumn{3}{c}{ACDC} &CMR T1-Map & \multirow{2}{*}{Fluoroscopy Image} \\
\cline{3-5} \cline{6-6}
 & & RV & Myo & LV & Myo & \\
\hline
\multirow{2}{*}{Full Data} 
  & U-Net       & 0.8743 & 0.8785 & 0.9473 & 0.8700 & 0.6539 \\
  & SwinUNETR   & 0.7623 & 0.8346 & 0.9198 & 0.8769& 0.6058 \\
\hline
\multirow{3}{*}{Limited Data (50 slices)} 
  & U-Net       & 0.1996 & 0.4794 & 0.5938 & 0.7792 & 0.5760 \\
  & SwinUNETR   & 0.2792 & 0.4202 & 0.4542 & 0.8147 & 0.5558 \\
  & Ours        & \textbf{0.6729} & \textbf{0.7757} & \textbf{0.8472} & \textbf{0.8238} & \textbf{0.7029} \\
\hline
\end{tabular}
\label{tab:tab2}
\end{table*}

In contrast, our method, which leverages the foundation model, achieves higher DSCs than supervised model in the limited data setting. Notably, on the Fluoroscopy image datasets, our method with limited data outperforms the supervised methods trained on the full dataset by 5\%-10\%. For the ACDC and CMR T1-Map datasets, our method's performance is only about 10\% lower in DSC. This suggests that the segmentation priors learned by the foundation model SAM 2 are highly generalizable. With a small amount of data, the model can quickly adapt to different tasks and achieve comparable or even superior performance compared to supervised methods designed for specific segmentation tasks.

\subsection{Comparison with Other Few-shot Medical Image Segmentation Methods}
We compare our method with other state-of-the-art (SOTA) methods for few-shot medical image segmentation, including PA-Net~\cite{wang2019panet}, ALP-Net~\cite{ouyang2020self}, AD-Net~\cite{hansen2022anomaly}, Q-Net~\cite{shen2023q} and CAT-Net~\cite{lin2023few}. For this comparison, we use the Abd-MRI dataset~\cite{kavur2021chaos}, which consists of 20 abdominal MRI scans with annotations for the left kidney (LK), right kidney (RK), spleen (Spl), and liver (Liv). We follow the 1-way 1-shot scenario and experiment setting described in~\cite{lin2023few} with 5-fold cross-validation. However, since our method does not require any training, we directly apply it to the test set for each fold. The results are demonstrated in \Cref{tab:tab4}.

\begin{table}[h]
\centering

\caption{Comparison of segmentation performance on the Abd-MRI dataset~\cite{kavur2021chaos} between our method and other few-shot medical image segmentation methods.}
\begin{tabular}{lccccc}
\hline
Methods& LK & RK & Spl. & Liv. & Avg. \\ 
\hline
PA-Net~\cite{wang2019panet} & 0.4771 & 0.4795 & 0.5873 & 0.6499 & 0.5485 \\ 
ALP-Net~\cite{ouyang2020self}&0.7363 & 0.7839 & 0.6702 & 0.7305 & 0.7302 \\ 
AD-Net~\cite{hansen2022anomaly}&0.7189 & 0.7602 & 0.6584 & 0.7603 & 0.7270 \\ 
Q-Net~\cite{shen2023q}&0.7405 & 0.7752 & 0.6743 & 0.7871 & 0.7443 \\ 
CAT-Net~\cite{lin2023few}&0.7401 & 0.7890 & 0.6883 & 0.7898 & 0.7518 \\ 
Ours& \textbf{0.7779}& \textbf{0.8581} & \textbf{0.7586} & \textbf{0.8793}&\textbf{0.8185}\\
\hline
\end{tabular}
\label{tab:tab4}

\end{table}

Our method achieves the highest DSCs across all evaluated organs. Specifically, our method obtains DSC of 0.7779 for the left kidney (LK), 0.8581 for the right kidney (RK), 0.7586 for the spleen (Spl.), and 0.8793 for the liver (Liv.), resulting in an average DSC of 0.8185. This outperforms the best-performing SOTA method, CAT-Net, which achieves an average DSC of 0.7518. The improvement is particularly notable for the spleen and liver, where our method surpasses CAT-Net by 7.03\% and 8.95\%, respectively. These results demonstrate the superior generalizability and adaptability of our method in few-shot medical image segmentation tasks. Notably, unlike those dedicated few-shot segmentation approaches, our method does not require any pretraining or training on base classes before testing on novel classes within the same data modality. This flexibility makes it well-suited for real-world clinical applications, where annotated data is often scarce, but the target regions and data modalities are diverse.

\subsection{Ablation Studies}

In this subsection, we conduct ablation studies to evaluate the impact of different retrieval strategies and the number of queried images on segmentation performance. 

\Cref{tab:tab3} compares the segmentation performance on the ACDC dataset using two retrieval strategies: Random and DINOv2, with two different numbers of queried images, \#8 and \#16.

\begin{table}[h!]
\centering
\caption{Comparison of segmentation performance on the ACDC dataset using different retrieval strategies (Random vs. DINOv2) and varying numbers of queried images (\#8 vs. \#16).}
\begin{tabular}{ccccc}
\hline
\multirow{2}{*}{No. Queried Imgs}& \multirow{2}{*}{Methods}& \multicolumn{3}{c}{ACDC} \\
\cline{3-5}
& & RV & Myo & LV \\
\hline
\multirow{2}{*}{\#8} & Random & 0.5801 & 0.7046 & 0.7796 \\
 & DINOv2 & 0.6529 & 0.7538 & 0.8183 \\
\multirow{2}{*}{\#16} & Random & 0.6684 & 0.7774 & 0.8402 \\
 & DINOv2 & 0.6729 & 0.7757 & 0.8472 \\
\hline
\end{tabular}
\label{tab:tab3}
\end{table}

In the case of retrieving 8 images (\#8), the DINOv2-based retrieval significantly outperforms the Random retrieval across all three classes. This indicates that the semantic information provided by the DINOv2 model is effective in selecting more relevant and similar support images, leading to better segmentation results. When the number of queried images is increased to 16 (\#16), both methods show improved performance, but the gap between the Random and DINOv2 methods narrows. This convergence can be attributed to the limited database size of 50 images. As the number of queried images increases, even randomly selected samples are likely to include some relevant examples, reducing the advantage of the DINOv2-based retrieval.

\begin{figure}[t]
\begin{center}
\includegraphics[width=1\linewidth]{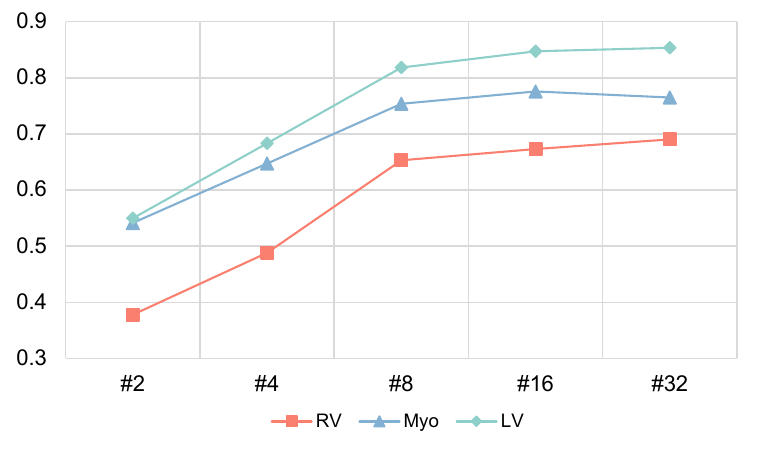}
\end{center}
\caption{Segmentation performance on the ACDC dataset with different numbers of queried images, measured by Dice similarity coefficient (DSC) for the right ventricle (RV), myocardium (Myo), and left ventricle (LV).}
\label{fig:fig4}
\end{figure}

We also explore the impact of the number of queried images on the segmentation performance for the ACDC dataset. The results are shown in \Cref{fig:fig4}. It is observed that as the number of queried images increases from 2 to 32, there is a consistent improvement in segmentation performance across all three classes. Initially, with only 2 queried images, the DSCs are relatively low for all classes, particularly for the RV. However, as the number of queried images increases to 4 and 8, there is a notable improvement in DSCs, with the performance curve showing a steep ascent, highlighting the importance of having a sufficient number of relevant support images to enhance segmentation accuracy. Beyond 8 queried images, the rate of improvement starts to plateau. This trend indicates that while increasing the number of queried images contributes to better segmentation performance, there is a diminishing return after a certain point, where additional images provide marginal gains.

\section{Discussion}

\subsection{Annotation Tools}
Our approach shows significant promise as an annotation tool for medical image segmentation, particularly in real-world practices where segmentation is often performed slice by slice. Traditional supervised methods often falter with limited annotated samples, making them less effective in such scenarios. However, our model can accurately segment challenging structures, like vessels which is time-consuming for human annotators, with limited data. Given just a few dozen samples, our model can quickly segment hundreds or thousands of images, producing the labels that can be used to train more robust supervised models. This efficiency can significantly accelerate the annotation process and enhance the overall quality of medical image segmentation.

Our model’s adaptability across different imaging modalities is a significant advantage, especially given the wide variation in medical images, such as differences in modality and contrast. Traditional models often struggle with this diversity, requiring extensive retraining for each new scenario. In contrast, our approach easily adapts to different imaging modalities by simply providing sample images and corresponding masks. This capability allows our model to effectively handle the heterogeneity of medical images, making it a versatile annotation tool for medical image segmentation across various clinical applications.

\subsection{Extend to 3D Medical Image Segmentation}

Our current method can be directly extend to 3D medical image segmentation by treating the 3D volume as a series of 2D slices. The database can store all slices from 3D volume of a subject or multiple subjects, and when a new subject is introduced, the model can segment the 3D volume slice by slice, retrieving the most similar slices from the database to guide the segmentation. For future work, performance could be further enhanced by adapting SAM 2 for 3D segmentation and incorporating specialized adaptors to capture volumetric information and spatial dependencies across slices.

\subsection{Running Time Analysis}
We have conducted a runtime comparison to assess the computational cost of the retrieval and memory-attention steps. Specifically, the DINOv2 encoding and retrieval process in our method takes an averaged 8.78 ms per slice with one NVIDIA A100 GPU, and the complete segmentation using our method requires 52.96/170.94/515.2 ms per slice when having 1/4/16 retrieved images to guide the segmentation. In comparison, SAM 2 with 4 point prompts takes 9.987 ms per slice. Although the increased runtime may limit real-time performance for high-throughput batch processing, it remains practical for clinical scenarios where accuracy and adaptability to limited data are more critical than processing speed. While SAM 2 with point prompts requires less time, its segmentation performance is not sufficient for clinical applications or to effectively reduce the annotation burden. In contrast, the segmentation results from our few-shot method can be used to train supervised segmentation models, helping to reduce the need for manual annotations or potentially being adapted for real-time processing with further optimization.

\subsection{Influence of Database Size}

We expand the database to include the full ACDC dataset (1808 images) for retrieval to invesitage the influence of including more samples in the database. We found that as the database size increases, the performance gain is minimal. The DSC for the right ventricle (RV) changed slightly from 0.6729 to 0.6695, for the myocardium (Myo) from 0.7757 to 0.7772, and for the left ventricle (LV) from 0.8472 to 0.8693. These results suggest that our model does not directly benefit from a larger database size, indicating that simply adding more diverse samples may not be the key to improving performance.

To further investigate the factors driving model performance, we conducted an experiment where we included the ground truth mask and the image to be segmented into the memory bank. This led to a remarkable improvement: the DSC for the right ventricle (RV) increased from 0.6729 to 0.9281, for the myocardium (Myo) from 0.7757 to 0.9075, and for the left ventricle (LV) from 0.8472 to 0.9577. This indicates that the similarity between the queried images and the test image is crucial for improving segmentation performance, as more similar images can provide more relevant and contextually aligned information for the model.

\subsection{Limitations}

{\textbf{FOV Variations.} In our experiments, the test image and the retrieved query images from the database have similar FOV and content, which enables DINOv2 to effectively measure global similarity. However, a significant FOV mismatch between the test and query images could limit the proposed method's applicability in annotation tools, such as when the database contains a full-body MRI image while the test image only contains a specific region like the heart. A potential solution could be to leverage the ground truth segmentation masks of full-body MRI images to crop them into smaller patches of varying sizes, each containing a specific organ or a combination of organs. This would increase the diversity of the database and enhance the model's ability to handle variations in FOV and content.

\textbf{Small Region Segmentaiton.} \Cref{fig:fig5} illustrates the segmentation performance of our method on challenging basal and apical slices from the ACDC dataset. Despite our method's overall effectiveness demonstrated in previous sections, the segmentation results for these particular slices are suboptimal. This is a common challenge even for supervised methods, as the basal and apical slices often present more anatomical variations and smaller regions of interest, making accurate segmentation difficult. The queried images visualized indicate that the retrieved samples, while semantically similar, may not provide adequate support due to the small size of the target regions and the inherent heterogeneity in medical images. To quantitatively investigate this influence, we computed the DSC separately for regions smaller than 200 pixels and those larger than 200 pixels in ACDC dataset for each class. We found that segmentation performance of our method varies depending on the size of the target region. For the right ventricle (RV), the DSC is 0.2723 for small regions (\textless 200 pixels) and increases to 0.8289 for large regions ($\geqslant$200 pixels). Similarly, for the myocardium (Myo), the DSC is 0.2226 for small regions and improves significantly to 0.8148 for large regions. For the left ventricle (LV), the DSC is 0.3402 for small regions and rises to 0.8814 for large regions. These findings align with the qualitative results shown \Cref{fig:fig5}. This suggests that the retrieval process in our method, which relies on whole-image features, may not always capture the fine details required for accurate segmentation of small objects, which needs future efforts.

\begin{figure}[h!]
\begin{center}
\includegraphics[width=1\linewidth]{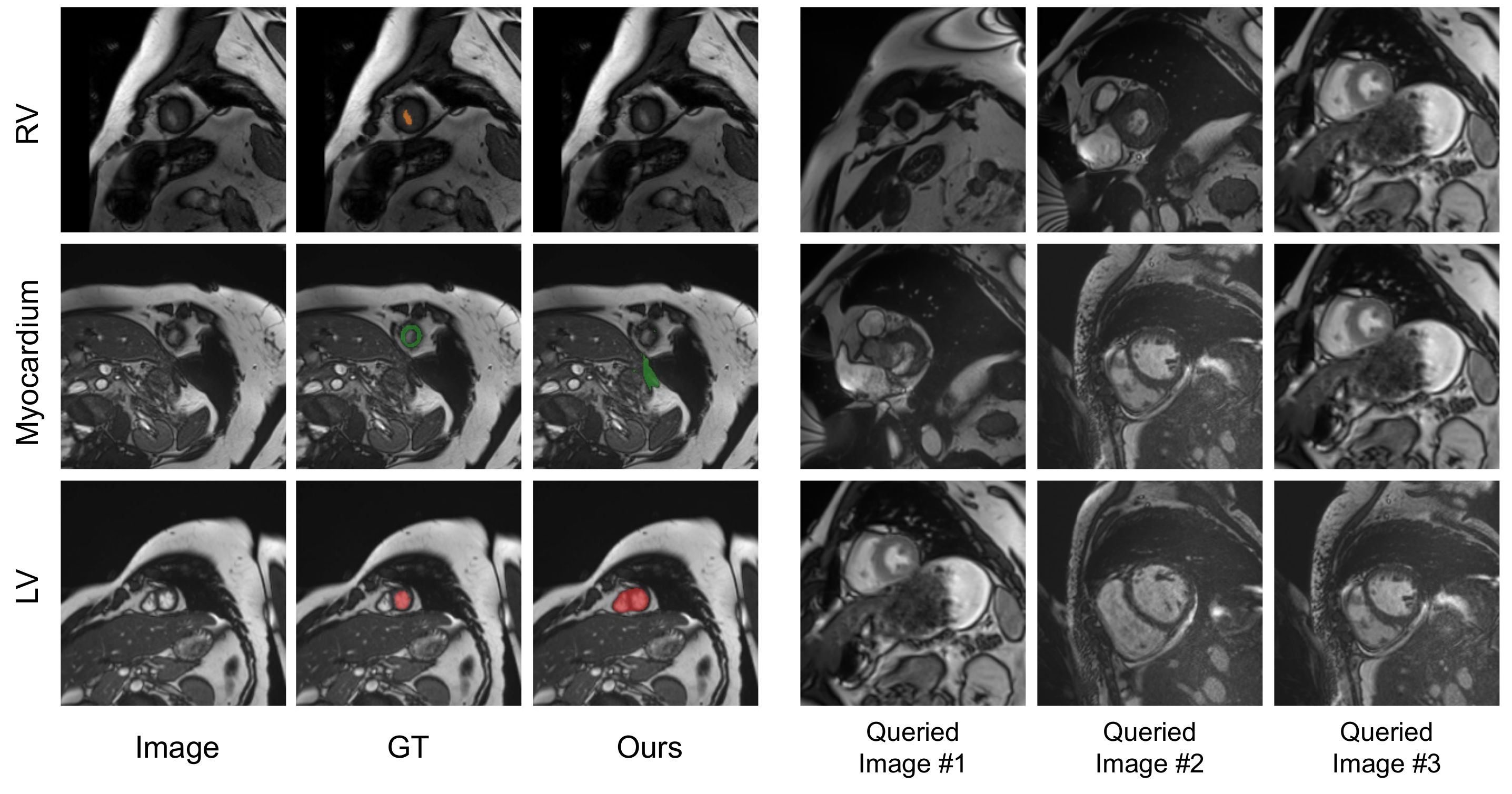}
\end{center}
\caption{Segmentation results on challenging basal and apical slices from the ACDC dataset, with corresponding top three similar queried images retrieved using the DINOv2.}
\label{fig:fig5}
\end{figure}

For the vessel segmentation task, we also observed that the model struggles when the image contrast is low, leading to suboptimal performance. This limitation stems from the inherent capabilities of the SAM 2 model, which our method builds upon. SAM 2 tends to underperform when the contrast between foreground/background is insufficient, especially for segmenting small branches in our cases (\Cref{fig:fig3}). These low-contrast vessels are particularly challenging to delineate accurately. One potential solution to this problem could be fine-tuning SAM 2 on a large-scale medical dataset, which might enhance its ability to handle low-contrast regions and capture the fine details necessary for accurate segmentation in these challenging scenarios.

\section{Conclusion}

In this paper, we introduced a retrieval-augmented few-shot segmentation framework for medical image segmentation, leveraging the advanced capabilities of foundation models like SAM 2 and DINOv2. Our approach consistently outperforms SAM 2 with point prompts, demonstrating improvements in segmentation accuracy and efficiency. Additionally, it shows remarkable adaptability across various medical imaging modalities, effectively handling the heterogeneity of medical data without the need for further retraining or finetuning. This method not only serves as a powerful tool for segmentation tasks but also holds great promise as a valuable annotation tool in clinical applications.

\bibliographystyle{IEEEtran}
\bibliography{reference}

\begin{thebibliography}{10}
\providecommand{\url}[1]{#1}
\csname url@samestyle\endcsname
\providecommand{\newblock}{\relax}
\providecommand{\bibinfo}[2]{#2}
\providecommand{\BIBentrySTDinterwordspacing}{\spaceskip=0pt\relax}
\providecommand{\BIBentryALTinterwordstretchfactor}{4}
\providecommand{\BIBentryALTinterwordspacing}{\spaceskip=\fontdimen2\font plus
\BIBentryALTinterwordstretchfactor\fontdimen3\font minus \fontdimen4\font\relax}
\providecommand{\BIBforeignlanguage}[2]{{%
\expandafter\ifx\csname l@#1\endcsname\relax
\typeout{** WARNING: IEEEtran.bst: No hyphenation pattern has been}%
\typeout{** loaded for the language `#1'. Using the pattern for}%
\typeout{** the default language instead.}%
\else
\language=\csname l@#1\endcsname
\fi
#2}}
\providecommand{\BIBdecl}{\relax}
\BIBdecl

\bibitem{rogowska2000overview}
J.~Rogowska, ``Overview and fundamentals of medical image segmentation,'' \emph{Handbook of medical imaging, processing and analysis}, pp. 69--85, 2000.

\bibitem{siddique2021u}
N.~Siddique, S.~Paheding, C.~P. Elkin, and V.~Devabhaktuni, ``U-net and its variants for medical image segmentation: A review of theory and applications,'' \emph{IEEE access}, vol.~9, pp. 82\,031--82\,057, 2021.

\bibitem{azad2024medical}
R.~Azad, E.~K. Aghdam, A.~Rauland, Y.~Jia, A.~H. Avval, A.~Bozorgpour, S.~Karimijafarbigloo, J.~P. Cohen, E.~Adeli, and D.~Merhof, ``Medical image segmentation review: The success of u-net,'' \emph{IEEE Transactions on Pattern Analysis and Machine Intelligence}, 2024.

\bibitem{chi2020deep}
W.~Chi, L.~Ma, J.~Wu, M.~Chen, W.~Lu, and X.~Gu, ``Deep learning-based medical image segmentation with limited labels,'' \emph{Physics in Medicine \& Biology}, vol.~65, no.~23, p. 235001, 2020.

\bibitem{cui2020unified}
H.~Cui, D.~Wei, K.~Ma, S.~Gu, and Y.~Zheng, ``A unified framework for generalized low-shot medical image segmentation with scarce data,'' \emph{IEEE Transactions on Medical Imaging}, vol.~40, no.~10, pp. 2656--2671, 2020.

\bibitem{peng2021medical}
J.~Peng and Y.~Wang, ``Medical image segmentation with limited supervision: a review of deep network models,'' \emph{IEEE Access}, vol.~9, pp. 36\,827--36\,851, 2021.

\bibitem{wang2019panet}
K.~Wang, J.~H. Liew, Y.~Zou, D.~Zhou, and J.~Feng, ``Panet: Few-shot image semantic segmentation with prototype alignment,'' in \emph{proceedings of the IEEE/CVF international conference on computer vision}, 2019, pp. 9197--9206.

\bibitem{ouyang2020self}
C.~Ouyang, C.~Biffi, C.~Chen, T.~Kart, H.~Qiu, and D.~Rueckert, ``Self-supervision with superpixels: Training few-shot medical image segmentation without annotation,'' in \emph{Computer Vision--ECCV 2020: 16th European Conference, Glasgow, UK, August 23--28, 2020, Proceedings, Part XXIX 16}.\hskip 1em plus 0.5em minus 0.4em\relax Springer, 2020, pp. 762--780.

\bibitem{hansen2022anomaly}
S.~Hansen, S.~Gautam, R.~Jenssen, and M.~Kampffmeyer, ``Anomaly detection-inspired few-shot medical image segmentation through self-supervision with supervoxels,'' \emph{Medical Image Analysis}, vol.~78, p. 102385, 2022.

\bibitem{shen2023q}
Q.~Shen, Y.~Li, J.~Jin, and B.~Liu, ``Q-net: Query-informed few-shot medical image segmentation,'' in \emph{Proceedings of SAI Intelligent Systems Conference}.\hskip 1em plus 0.5em minus 0.4em\relax Springer, 2023, pp. 610--628.

\bibitem{lin2023few}
Y.~Lin, Y.~Chen, K.-T. Cheng, and H.~Chen, ``Few shot medical image segmentation with cross attention transformer,'' in \emph{International Conference on Medical Image Computing and Computer-Assisted Intervention}.\hskip 1em plus 0.5em minus 0.4em\relax Springer, 2023, pp. 233--243.

\bibitem{ma2024segment}
J.~Ma, Y.~He, F.~Li, L.~Han, C.~You, and B.~Wang, ``Segment anything in medical images,'' \emph{Nature Communications}, vol.~15, no.~1, p. 654, 2024.

\bibitem{wu2023medical}
J.~Wu, W.~Ji, Y.~Liu, H.~Fu, M.~Xu, Y.~Xu, and Y.~Jin, ``Medical sam adapter: Adapting segment anything model for medical image segmentation,'' \emph{arXiv preprint arXiv:2304.12620}, 2023.

\bibitem{lei2023medlsam}
W.~Lei, X.~Wei, X.~Zhang, K.~Li, and S.~Zhang, ``Medlsam: Localize and segment anything model for 3d medical images,'' \emph{arXiv preprint arXiv:2306.14752}, 2023.

\bibitem{bui2023sam3d}
N.-T. Bui, D.-H. Hoang, M.-T. Tran, and N.~Le, ``Sam3d: Segment anything model in volumetric medical images,'' \emph{arXiv preprint arXiv:2309.03493}, 2023.

\bibitem{ronneberger2015u}
O.~Ronneberger, P.~Fischer, and T.~Brox, ``U-net: Convolutional networks for biomedical image segmentation,'' in \emph{Medical image computing and computer-assisted intervention--MICCAI 2015: 18th international conference, Munich, Germany, October 5-9, 2015, proceedings, part III 18}.\hskip 1em plus 0.5em minus 0.4em\relax Springer, 2015, pp. 234--241.

\bibitem{isensee2021nnu}
F.~Isensee, P.~F. Jaeger, S.~A. Kohl, J.~Petersen, and K.~H. Maier-Hein, ``nnu-net: a self-configuring method for deep learning-based biomedical image segmentation,'' \emph{Nature methods}, vol.~18, no.~2, pp. 203--211, 2021.

\bibitem{ravi2024sam}
N.~Ravi, V.~Gabeur, Y.-T. Hu, R.~Hu, C.~Ryali, T.~Ma, H.~Khedr, R.~R{\"a}dle, C.~Rolland, L.~Gustafson \emph{et~al.}, ``Sam 2: Segment anything in images and videos,'' \emph{arXiv preprint arXiv:2408.00714}, 2024.

\bibitem{lewis2020retrieval}
P.~Lewis, E.~Perez, A.~Piktus, F.~Petroni, V.~Karpukhin, N.~Goyal, H.~K{\"u}ttler, M.~Lewis, W.-t. Yih, T.~Rockt{\"a}schel \emph{et~al.}, ``Retrieval-augmented generation for knowledge-intensive nlp tasks,'' \emph{Advances in Neural Information Processing Systems}, vol.~33, pp. 9459--9474, 2020.

\bibitem{edge2024local}
D.~Edge, H.~Trinh, N.~Cheng, J.~Bradley, A.~Chao, A.~Mody, S.~Truitt, and J.~Larson, ``From local to global: A graph rag approach to query-focused summarization,'' \emph{arXiv preprint arXiv:2404.16130}, 2024.

\bibitem{oquab2023dinov2}
M.~Oquab, T.~Darcet, T.~Moutakanni, H.~Vo, M.~Szafraniec, V.~Khalidov, P.~Fernandez, D.~Haziza, F.~Massa, A.~El-Nouby \emph{et~al.}, ``Dinov2: Learning robust visual features without supervision,'' \emph{arXiv preprint arXiv:2304.07193}, 2023.

\bibitem{bernard2018deep}
O.~Bernard, A.~Lalande, C.~Zotti, F.~Cervenansky, X.~Yang, P.-A. Heng, I.~Cetin, K.~Lekadir, O.~Camara, M.~A.~G. Ballester \emph{et~al.}, ``Deep learning techniques for automatic mri cardiac multi-structures segmentation and diagnosis: is the problem solved?'' \emph{IEEE transactions on medical imaging}, vol.~37, no.~11, pp. 2514--2525, 2018.

\bibitem{kavur2021chaos}
A.~E. Kavur, N.~S. Gezer, M.~Bar{\i}{\c{s}}, S.~Aslan, P.-H. Conze, V.~Groza, D.~D. Pham, S.~Chatterjee, P.~Ernst, S.~{\"O}zkan \emph{et~al.}, ``Chaos challenge-combined (ct-mr) healthy abdominal organ segmentation,'' \emph{Medical image analysis}, vol.~69, p. 101950, 2021.

\bibitem{radford2018improving}
A.~Radford, K.~Narasimhan, T.~Salimans, I.~Sutskever \emph{et~al.}, ``Improving language understanding by generative pre-training,'' 2018.

\bibitem{achiam2023gpt}
J.~Achiam, S.~Adler, S.~Agarwal, L.~Ahmad, I.~Akkaya, F.~L. Aleman, D.~Almeida, J.~Altenschmidt, S.~Altman, S.~Anadkat \emph{et~al.}, ``Gpt-4 technical report,'' \emph{arXiv preprint arXiv:2303.08774}, 2023.

\bibitem{dubey2024llama}
A.~Dubey, A.~Jauhri, A.~Pandey, A.~Kadian, A.~Al-Dahle, A.~Letman, A.~Mathur, A.~Schelten, A.~Yang, A.~Fan \emph{et~al.}, ``The llama 3 herd of models,'' \emph{arXiv preprint arXiv:2407.21783}, 2024.

\bibitem{caron2021emerging}
M.~Caron, H.~Touvron, I.~Misra, H.~J\'egou, J.~Mairal, P.~Bojanowski, and A.~Joulin, ``Emerging properties in self-supervised vision transformers,'' in \emph{Proceedings of the International Conference on Computer Vision (ICCV)}, 2021.

\bibitem{radford2021learning}
A.~Radford, J.~W. Kim, C.~Hallacy, A.~Ramesh, G.~Goh, S.~Agarwal, G.~Sastry, A.~Askell, P.~Mishkin, J.~Clark \emph{et~al.}, ``Learning transferable visual models from natural language supervision,'' in \emph{International conference on machine learning}.\hskip 1em plus 0.5em minus 0.4em\relax PMLR, 2021, pp. 8748--8763.

\bibitem{liu2024visual}
H.~Liu, C.~Li, Q.~Wu, and Y.~J. Lee, ``Visual instruction tuning,'' \emph{Advances in neural information processing systems}, vol.~36, 2024.

\bibitem{tumanyan2024dino}
N.~Tumanyan, A.~Singer, S.~Bagon, and T.~Dekel, ``Dino-tracker: Taming dino for self-supervised point tracking in a single video,'' \emph{arXiv preprint arXiv:2403.14548}, 2024.

\bibitem{jiang2024omniglue}
H.~Jiang, A.~Karpur, B.~Cao, Q.~Huang, and A.~Araujo, ``Omniglue: Generalizable feature matching with foundation model guidance,'' in \emph{Proceedings of the IEEE/CVF Conference on Computer Vision and Pattern Recognition}, 2024, pp. 19\,865--19\,875.

\bibitem{kirillov2023segment}
A.~Kirillov, E.~Mintun, N.~Ravi, H.~Mao, C.~Rolland, L.~Gustafson, T.~Xiao, S.~Whitehead, A.~C. Berg, W.-Y. Lo \emph{et~al.}, ``Segment anything,'' in \emph{Proceedings of the IEEE/CVF International Conference on Computer Vision}, 2023, pp. 4015--4026.

\bibitem{mazurowski2023segment}
M.~A. Mazurowski, H.~Dong, H.~Gu, J.~Yang, N.~Konz, and Y.~Zhang, ``Segment anything model for medical image analysis: an experimental study,'' \emph{Medical Image Analysis}, vol.~89, p. 102918, 2023.

\bibitem{blattmann2022retrieval}
A.~Blattmann, R.~Rombach, K.~Oktay, J.~M{\"u}ller, and B.~Ommer, ``Retrieval-augmented diffusion models,'' \emph{Advances in Neural Information Processing Systems}, vol.~35, pp. 15\,309--15\,324, 2022.

\bibitem{long2022retrieval}
A.~Long, W.~Yin, T.~Ajanthan, V.~Nguyen, P.~Purkait, R.~Garg, A.~Blair, C.~Shen, and A.~van~den Hengel, ``Retrieval augmented classification for long-tail visual recognition,'' in \emph{Proceedings of the IEEE/CVF conference on computer vision and pattern recognition}, 2022, pp. 6959--6969.

\bibitem{liu2023learning}
H.~Liu, K.~Son, J.~Yang, C.~Liu, J.~Gao, Y.~J. Lee, and C.~Li, ``Learning customized visual models with retrieval-augmented knowledge,'' in \emph{Proceedings of the IEEE/CVF Conference on Computer Vision and Pattern Recognition}, 2023, pp. 15\,148--15\,158.

\bibitem{douze2024faiss}
M.~Douze, A.~Guzhva, C.~Deng, J.~Johnson, G.~Szilvasy, P.-E. Mazaré, M.~Lomeli, L.~Hosseini, and H.~Jégou, ``The faiss library,'' 2024.

\bibitem{darcet2023vitneedreg}
T.~Darcet, M.~Oquab, J.~Mairal, and P.~Bojanowski, ``Vision transformers need registers,'' 2023.

\bibitem{ryali2023hiera}
C.~Ryali, Y.-T. Hu, D.~Bolya, C.~Wei, H.~Fan, P.-Y. Huang, V.~Aggarwal, A.~Chowdhury, O.~Poursaeed, J.~Hoffman \emph{et~al.}, ``Hiera: A hierarchical vision transformer without the bells-and-whistles,'' in \emph{International Conference on Machine Learning}.\hskip 1em plus 0.5em minus 0.4em\relax PMLR, 2023, pp. 29\,441--29\,454.

\bibitem{hatamizadeh2021swin}
A.~Hatamizadeh, V.~Nath, Y.~Tang, D.~Yang, H.~R. Roth, and D.~Xu, ``Swin unetr: Swin transformers for semantic segmentation of brain tumors in mri images,'' in \emph{International MICCAI brainlesion workshop}.\hskip 1em plus 0.5em minus 0.4em\relax Springer, 2021, pp. 272--284.

\end{thebibliography}

\end{document}